\documentclass[letterpaper, 10 pt, conference]{ieeeconf}  % Comment this line out if you need a4paper

\IEEEoverridecommandlockouts                              % This command is only needed if 
\usepackage{graphics} % for pdf, bitmapped graphics files
\usepackage{epsfig} % for postscript graphics files
\usepackage{mathptmx} % assumes new font selection scheme installed
\usepackage{times} % assumes new font selection scheme installed
\usepackage{amsmath} % assumes amsmath package installed
\usepackage{amssymb}  % assumes amsmath package installed
\usepackage{graphicx}
\usepackage{subfig}
\usepackage{multirow}
\usepackage{booktabs}
\usepackage{bm}
\usepackage{hyperref}
\usepackage{cleveref}
\usepackage{makecell}
\usepackage{pifont}

\DeclareMathOperator*{\argmin}{arg\,min}
\newcommand{\cmark}{\ding{51}}%
\newcommand{\xmark}{\ding{55}}%

\title{\LARGE 
\textbf{SoftMAC}: Differentiable \textbf{Soft} Body Simulation with Forecast-based Contact \textbf{M}odel
and Two-way Coupling with \textbf{A}rticulated Rigid Bodies and \textbf{C}lothes
}

\author{Min Liu$^{1}$, Gang Yang$^{2}$, Siyuan Luo$^{2}$, and Lin Shao$^{2}$% <-this % stops a space
\thanks{$^{1}$Min Liu is with Machine Learning Department, Carnegie Mellon University, USA. 
        {\tt\small minliu2@cs.cmu.edu}}%
\thanks{$^{2}$Gang Yang, Siyuan Luo and Lin Shao are with Department of Computer Science, National University of Singapore, Singapore.
        {\tt\small yg.matinal@gmail.com, luosiyuan2002@gmail.com and linshao@nus.edu.sg}}%
}

\begin{document}

\maketitle
\thispagestyle{empty}
\pagestyle{empty}

%%%%%%%%%%%%%%%%%%%%%%%%%%%%%%%%%%%%%%%%%%%%%%%%%%%%%%%%%%%%%%%%%%%%%%%%%%%%%%%%
\begin{abstract}
Differentiable physics simulation provides an avenue to tackle previously intractable challenges through gradient-based optimization, thereby greatly improving the efficiency of solving robotics-related problems.
To apply differentiable simulation in diverse robotic manipulation scenarios, a key challenge is to integrate various materials in a unified framework. 
We present SoftMAC, a differentiable simulation framework that couples soft bodies with articulated rigid bodies and clothes. 
SoftMAC simulates soft bodies with the continuum-mechanics-based Material Point Method (MPM). We provide a novel forecast-based contact model for MPM, which effectively reduces penetration without introducing other artifacts like unnatural rebound. 
To couple MPM particles with deformable and non-volumetric clothes meshes, we also propose a penetration tracing algorithm that reconstructs the signed distance field in local area.
Diverging from previous works, SoftMAC simulates the complete dynamics of each modality and incorporates them into a cohesive system with an explicit and differentiable coupling mechanism.
The feature empowers SoftMAC to handle a broader spectrum of interactions, such as soft bodies serving as manipulators and engaging with underactuated systems.
We conducted comprehensive experiments to validate the effectiveness and accuracy of the proposed differentiable pipeline in downstream robotic manipulation applications. Supplementary materials are available on our project website at \href{https://minliu01.github.io/SoftMAC/}{https://minliu01.github.io/SoftMAC}.
\end{abstract}

%%%%%%%%%%%%%%%%%%%%%%%%%%%%%%%%%%%%%%%%%%%%%%%%%%%%%%%%%%%%%%%%%%%%%%%%%%%%%%%%
\section{INTRODUCTION}
\begin{figure}[t]
    \centering
    \includegraphics[width=0.91\linewidth]{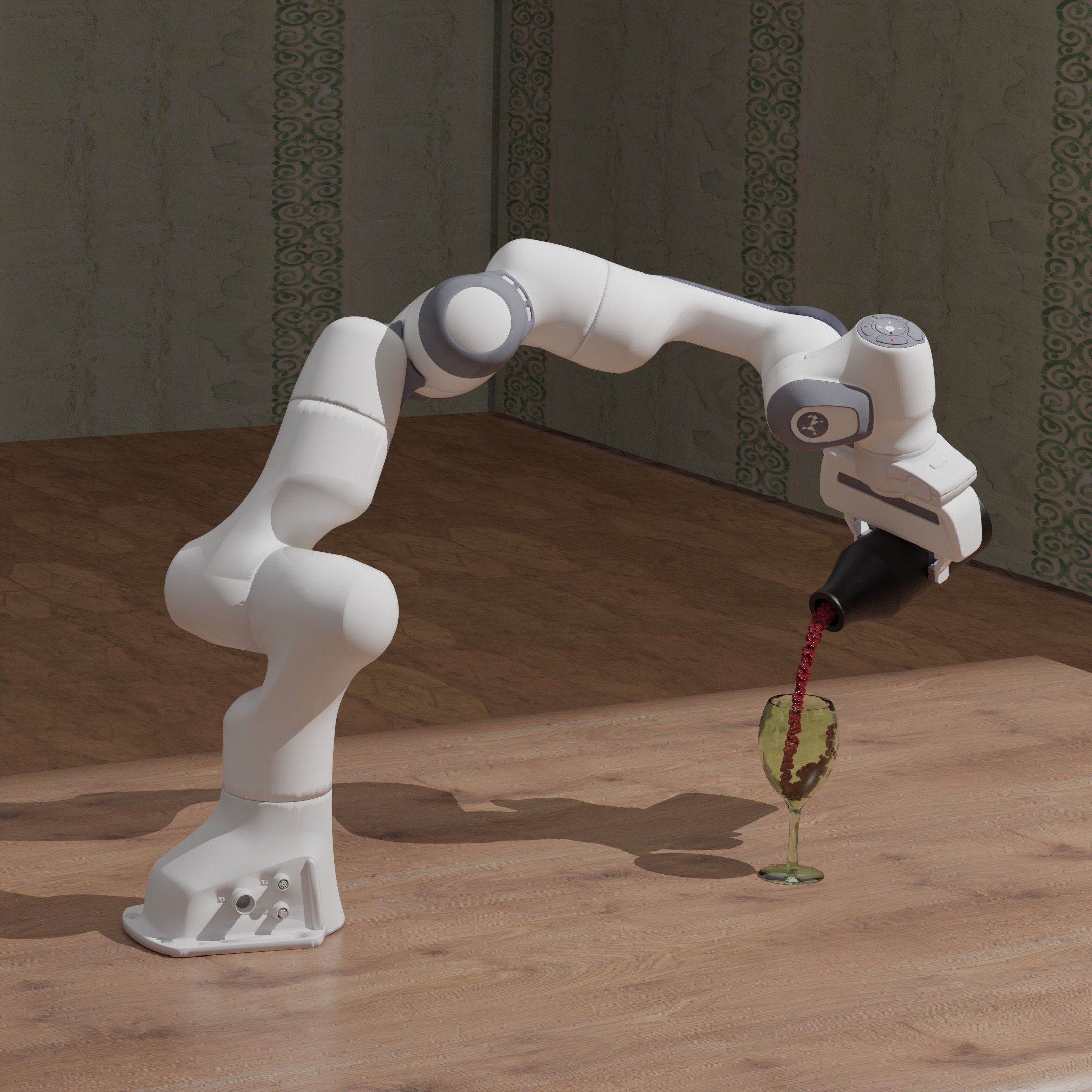}
    \caption{
        To pour water into a glass in differentiable physics simulation, we need realistic contact model for soft-rigid coupling and correct gradient calculation. Then same applies to the interactions between soft bodies and clothes.
    }
    \label{fig:teaser}
\end{figure}

Interactions between diverse materials are prevalent in the realm of robotic manipulation. 
For instance, the collision between glass and wine during a pouring task as in \cref{fig:teaser}, or the perpetual interplay between a tortilla and its fillings when crafting a taco. 
As a number of differentiable physics engines emerge to tackle learning and control problems \cite{werling2021fast, howell2022dojo, hu2019chainqueen, xian2023fluidlab, liang2019differentiable, li2022diffcloth}, it is tempting to unify different materials in a single simulator, thus supporting a wide range of manipulation tasks. 
Nevertheless, developing a general-purpose physics simulator is non-trivial, as different types of materials often need different physical models due to the dominant physical behaviors \cite{liang2020differentiable}.

Depending on their dynamical behaviors, most daily objects can be simulated as: 1) soft bodies, like elastic, plastic, elasto-plastic objects and liquid; 2) rigid bodies, either articulated or not; 3) clothes, as well as other thin-shell objects with similar geometric and physical properties (\textit{e.g.}, tortilla).
The interactions between these three modalities encompass a wide spectrum of manipulation tasks. 
Rigid-clothes coupling has been extensively discussed in previous works \cite{bai2014coupling,macklin2014unified,yu2023diffclothai}. 
% DiffClothAI \cite{yu2023diffclothai} provides intersection-free frictional contact and a differentiable pipeline to couple clothes with articulated rigid bodies.
Meanwhile, differentiable soft-cloth coupling remains a rarely explored topic in current literature, despite its prevalence in robotic manipulation scenarios such as making taco.
A line of works propose methods for soft-rigid coupling. 
PlasticineLab \cite{huang2021plasticinelab}, FluidLab \cite{xian2023fluidlab} and DexDeform \cite{li2023dexdeform} deploys Material Point Method (MPM) to simulate soft bodies, but they only simulate the kinematics of rigid bodies, implying that force exerted on soft bodies cannot react on rigid bodies. 
Maniskill2 \cite{gu2023maniskill2} introduces two-way dynamics coupling, but its pipeline is not differentiable. 
Besides, contact models for MPM in these works suffer from artifacts like penetration and unnatural rebound.

\begin{table*} [t!]
    \centering
    \setlength\tabcolsep{3.8pt}
    \renewcommand{\arraystretch}{1.15}
    \begin{tabular}{l|c|ccc|cc|c|ccc}
    \toprule[1.2pt]
        \multicolumn{1}{c|}{\multirow{2}{*}{\textbf{Simulator}}} & \multicolumn{1}{c|}{\multirow{2}{*}{\textbf{Differentiable}}} & \multicolumn{3}{c|}{{\textbf{\quad MPM\quad }}} & \multicolumn{2}{c|}{\textbf{Rigid}} & \multicolumn{1}{c|}{\multirow{2}{*}{\textbf{\quad Cloth\quad }}} & \multicolumn{3}{c}{\textbf{Coupling}} \\\cline{3-7}\cline{9-11}
         & & \textbf{Elastic} & \textbf{Plastic} & \textbf{Liquid} & \textbf{Dynamics} & \textbf{Articulated} & & \textbf{Contact} & \textbf{MPM-Rigid} & \textbf{MPM-Cloth}\\\hline
        \textbf{PlasticineLab \cite{huang2021plasticinelab}} & \cmark & & \cmark & & & & & Grid & One-way & $\backslash$ \\
        \textbf{FluidLab \cite{du2012fluid}} & \cmark & \cmark & \cmark & \cmark & * & & & Grid/Particle & One-way & $\backslash$ \\
        \textbf{DexDeform \cite{li2023dexdeform}}  & \cmark & & \cmark & & & $\dagger$ & & Grid & One-way & $\backslash$ \\
        \textbf{ManiSkill2 \cite{gu2023maniskill2}} & & \cmark & \cmark & \cmark &\cmark & \cmark & & Grid/Particle & Two-way & $\backslash$ \\
        \textbf{SoftMAC (Ours)} & \cmark & \cmark & \cmark & \cmark & \cmark & \cmark & \cmark & Forecast & Two-way & Two-way \\
    \bottomrule[1.2pt]
    \end{tabular}
    \caption{
        Comparison with other popular MPM-based simulators. 
        *FluidLab supports modeling rigid bodies using kinematic skeletons or MPM. While the latter can simulate dynamics of rigid bodies, it is limited to simple shapes without articulation, which is not considered as general-purpose rigid body simulation.
        $\dagger$DexDeform provides a simulator tailored for deformable object manipulation with the Shadow hand. The specialization limits its direct applicability to other articulated rigid bodies.
        % Description of 6 trajectory optimization tasks. \textit{\#Steps} denotes the length of action sequences. \textit{DoF} records the dimension of action and degree of freedom of the system (MPM + the other modality). \textit{Modality} displays the materials involved in the task and whether MPM soft body functions as the manipulator. \textit{Optimization} presents the optimizer, number of variables to optimize, and number of iterations.
    }
    \label{tab:simulation_comparison}
\end{table*}

% \looseness=-1
In this paper, we present a differentiable pipeline to couple soft bodies with articulated rigid bodies and clothes. 
Following previous works, we adopt MPM for soft bodies due to its ability to simulate a large variety of deformable materials and physical phenomena (e.g., Magnus effect and buoyancy) \cite{du2012fluid}. 
% Rigid bodies and clothes are modeled with existing simulators \cite{gang2023jade, yu2023diffclothai}, and represented as meshes in coupling. 
% We extend the capabilities of differentiable soft body simulators in the following two directions:
% \textbf{Forecase-based Contact Modeling.} 
MPM struggles with delicate boundaries. Scaling down grid size and time step alleviates the problem, but is not always feasible in robotic simulation where computational efficiency should be balanced. 
Gu \textit{et al.} \cite{gu2023maniskill2} apply particle forces to reduce penetration, but the method introduces problems such as unnatural rebound.
To achieve realistic collision effects, we introduce a novel method called \textit{forecast-based contact model} to manage the boundary conditions for MPM. 
Specifically, the model takes a grid-to-particle transfer to look ahead in the grid operation stage, imposes constraints on particles within the contact region, and then adjusts the grid velocity accordingly. 
Forecast-based contact model requires signed distance fields (SDFs) to penalize penetration. 
While the definition of SDF is straightforward for volumetric objects, it is hard to determine the sign on nonvolumetric meshes. 
To solve the problem, we propose a \textit{penetration tracing} algorithm that capitalizes on the localized motion of particles to reconstruct the SDF within confined zones. 
In this way, the contact model can be applied to both soft-rigid and soft-cloth coupling. 
% We show that the method effectively reduces artifacts like penetration in quantitative and qualitative experiments.

% \textbf{Two-way Differentiable Dynamics Coupling.}
Previous differentiable MPM simulators \cite{xian2023fluidlab,huang2021plasticinelab,li2023dexdeform} model external materials as kinematic skeletons, without explicitly considering forces, mass distribution, or other dynamic factors.
While it simplifies the pipeline, the simulators cannot generalize to scenarios where dynamics of both materials is involved (e.g., underactuated systems).
We overcome the problem by utilizing independent dynamics simulation for each modality and incorporating them into a cohesive system.
As in \cite{gu2023maniskill2, han2019hybrid}, we handle the contact between MPM and other materials explicitly, but make the whole process differentiable.
Specifically, we copy motion of rigid bodies and clothes to the soft body simulator, and transfer force back in forward simulation. 
Then we apply chain rule in reverse direction along the computation graph to propagate gradients between different simulators.
The gradient information allows us to control the motion between different modalities by optimizing the force acting on any point in the simulation. 
% As a result, we implement a differentiable simulator that enables manipulating soft bodies with articulated rigid bodies and clothes. 
Code of the proposed simulator, which we call SoftMAC, is released at our project website. 

Our main contributions can be summarized as follows:
% \vspace{-0.2em}
\begin{itemize}
\setlength\itemsep{0.0em}
    % \item We propose a contact model for MPM. Artifacts like penetration and unnatural rebound are greatly reduced.
    \item We propose a novel forecast-based contact model for MPM, which reduces penetration without introducing artifacts like unnatural rebound.
    % \item We present a differentiable soft body simulator that couples with articulated rigid bodies and clothes, and show its efficacy under several manipulation tasks.
    \item We present a penetration tracing algorithm for the contact between MPM and non-volumetric cloth meshes.
    \item To the best of our knowledge, SoftMAC is the first differentiable robotic simulator to support two-way dynamics soft-rigid and soft-cloth coupling. % Our work lays a foundation for universal differentiable physics simulation.
\end{itemize}

\section{Related Work}
\subsection{Differentiable Physics-based Simulation}
Physics simulation provides an avenue to avoid real-world damages, accelerate robotics-related problem solving, and establishes standardized benchmarks to evaluate different algorithms. 
A branch of works provide simulation for articulated rigid bodies \cite{todorov2012mujoco, coumans2016pybullet, koenig2004design, xiang2020sapien}, which is fundamental in robotics. 
With advances in deformable object manipulation, an increasing number of physics engines are also proposed for clothes and soft bodies \cite{lin2021softgym, gan2021threedworld}.
Recent developments in automatic differentiation methods \cite{paszke2019pytorch, bradbury2018jax, hu2019difftaichi} boost the prosperity in differentiable physics simulators, which can convert challenging control tasks into gradient-based optimization problems. 
Several differentiable simulators are developed for rigid bodies \cite{werling2021fast, howell2022dojo}, clothes \cite{li2022diffcloth, yu2023diffclothai, stuyck2023diffxpbd}, thin shells \cite{wang2024thinshell}, and soft bodies \cite{du2021_diffpd,huang2022differentiable}. 
A line of works \cite{hu2019chainqueen, xian2023fluidlab, huang2021plasticinelab} simulate soft bodies with material point method (MPM) to support various materials and physical processes. 
We also deploy MPM for soft bodies in this work.

\begin{figure*} [t!]
\centering
    \subfloat[Contact Models\label{fig:contact_models}]{
        \includegraphics[width=0.5\linewidth]{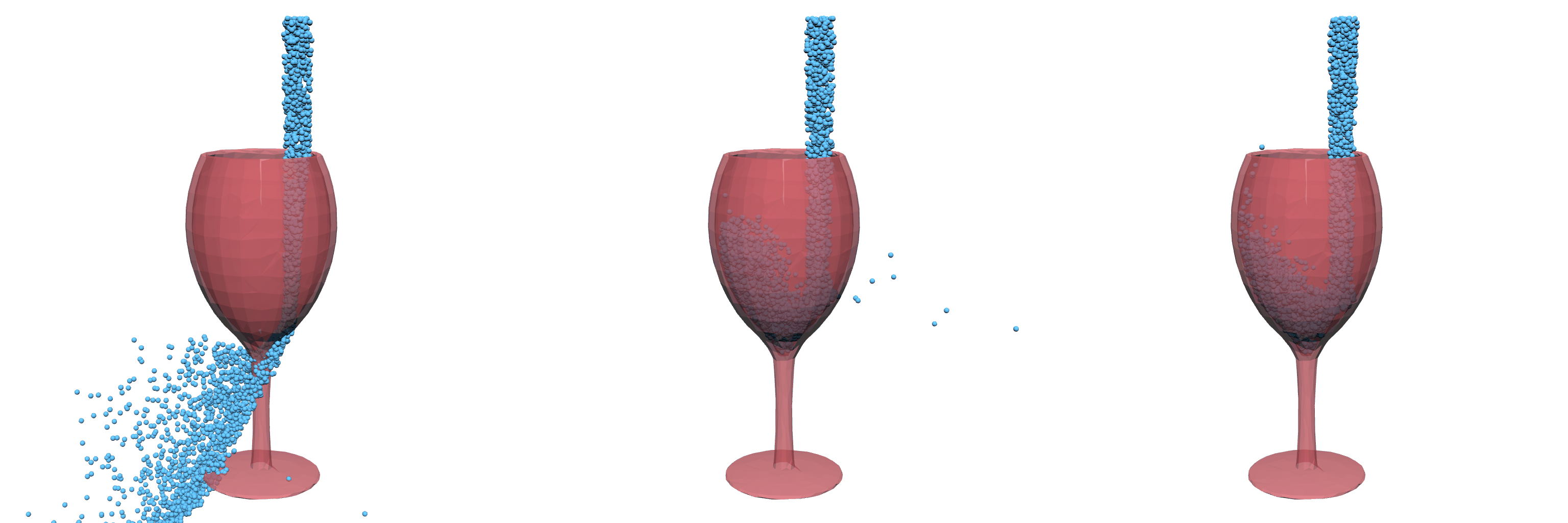}
    }
    \subfloat[Penetration Tracing\label{fig:penetration_tracing}]
    {
        \includegraphics[width=0.4\linewidth]{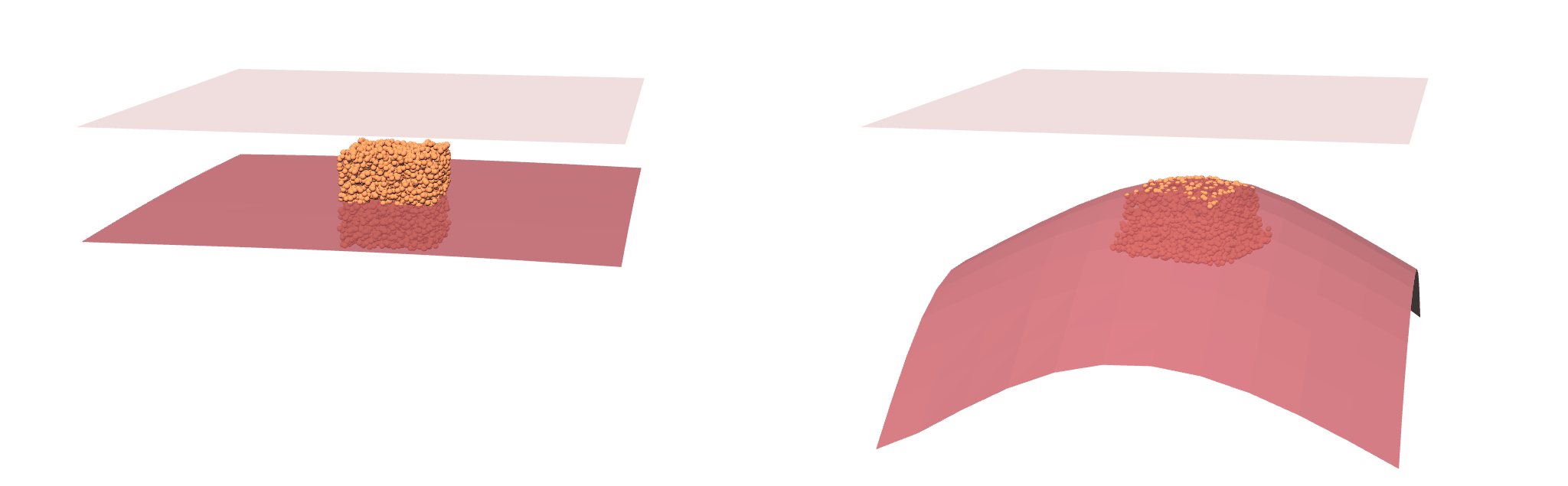}
    }
    \caption{
    \textit{(a)} Pour water into a thin glass. Grid-based model (left) leads to severe penetration.
    Particle-based model (middle) also causes a few particles to penetrate the glass.
    Forecast-based model (right) achieves the most robust performance.
    \textit{(b)} Drag four corners of a towel to squash a plasticine.
    Towel goes through the plasticine without penetration tracing (left).
    Both the plasticine and towel deform due to the contact after adding the algorithm (right).
    }
    \label{fig:contact_model_demos} 
\end{figure*}

\subsection{Coupling Soft Bodies with Other Modalities}
Unifying soft bodies with other modalities in a single framework requires the design of contact mechanisms between different materials. A number of methods have been proposed in robotics and computer graphics area.
Du \textit{et al.} \cite{du2012fluid} and Harada \textit{et al.} \cite{harada2007real} provide simulation methods to couple cloth and fluid computed by using smoothed
particle hydrodynamics.
MuJoCo \cite{todorov2012mujoco} and Bullet \cite{coumans2016pybullet} enable manipulating soft bodies simulated with finite element method. However, these methods only support a limited number of soft body materials.
Recent works based on MPM provide the opportunity to couple various types of soft bodies with other modalities.
PlasticineLab \cite{huang2021plasticinelab} and FluidLab \cite{du2012fluid} provide interaction with rigid bodies by penalizing grid velocity directly, but their contact models are one-way.
Maniskill2 \cite{gu2023maniskill2} transfers force back to rigid bodies to enable bidirectional contact, and penalize particles in their contact models to alleviate penetration. 
Nevertheless, their pipeline does not provide gradient information. 
Our work is based on a novel contact model better at reducing penetration. It also provides bidirectional contact for both soft-rigid and soft-cloth coupling, and make the process differentiable.
We present a comparison in terms of differentiability, material types, and coupling methods with other MPM-based simulators in \cref{tab:simulation_comparison}.
% SoftMAC covers a wider range of materials and interactions between them, enabling more diverse robotic manipulation scenarios.
SoftMAC covers a wider range of materials and their interactions, thereby facilitating a more extensive array of robotic manipulation scenarios.

\subsection{Robotic Manipulation with Differentiable Simulation}
Differentiable physics simulation has been applied in several robotic manipulation methods \cite{wang2024thinshell, xiang2023diff,murthy2020gradsim,lin2022diffskill}.
DexDeform \cite{li2023dexdeform} trains a skill model from human demonstrations, and uses a gradient optimizer to refine the trajectories planned by the skill model to generate more demonstrations.
SAGCI \cite{lv2022sagci} proposes a model-based learning method with differentiable simulation to online verify and modify the environment model during interaction.
SAM-RL \cite{lv2022sam} combines differentiable physics simulation and rendering to propose a sensing-aware learning pipeline that selects an informative viewpoint to monitor the manipulation process.

\section{Contact Models}

\subsection{Preliminary}
We build our soft body simulator based on MLS-MPM \cite{hu2018moving}. For simplicity, we use \textit{MPM} for \textit{MLS-MPM} in this section. 
In a simulation loop, the matter is interpolated back and forth between particle and grid representations. 
Consider a system that consists of $n_p$ particles and $n_g$ grid nodes. Denote $x_p, v_p, m_p \in \mathbb{R}^{3n_p}$ as positions, velocities and masses of particles, and $v_g, p_g, m_g \in \mathbb{R}^{3n_g}$ as velocities, momentum and masses of grid nodes.
We duplicate the masses to higher dimension for the convenience of vector operations.
$W: \mathbb{R}^{3n_p} \rightarrow \mathbb{R}^{3n_g \times 3n_p}$ calculates a sparse matrix that contains weights for the $3^3$ neighbouring grid nodes surrounding each particle.
Superscript denotes the time step of a variable.

\paragraph{Particle-to-Grid (P2G)} MPM first computes stress based on the constitutive model, and then transfer particles' momentum and masses to background grid using the interpolation matrix $W(x_p^n)$:
\begin{equation} 
\vspace{-2pt}
\begin{split}
    p_g^{n+1} &= W(x^{n}_p)(m_p^{\top} v_p^n + p_d^n), \\
    m_g^{n+1} &= W(x^{n}_p)m_p^{n}.
\end{split}
\label{eq:p2g}
\vspace{-2pt}
\end{equation}
where $p_d^n$ is a momentum term that reflects internal forces.

\paragraph{Grid Operation} The discrete equations of momentum are solved on grid nodes. Boundary conditions $BC(\cdot)$ can be enforced at this stage to set constraints on grid velocity:
\begin{equation}
\vspace{-2pt}
\begin{split}
    \hat{v}_g^{n+1} &= (p_g^{n+1})^{\top}(1 / m_g^{n+1}),\\
    v_g^{n+1} &= BC(\hat{v}_g^{n+1}).
\end{split}
\vspace{-2pt}
\label{eq:gridop}
\end{equation}

\paragraph{Grid-to-Particle (G2P)} The final stage transfers velocity back to particles, updates deformation information, and advects particles based on the new velocities.
\begin{equation}
\begin{split}
\vspace{-2pt}
    v_p^{n+1} &=  W( x_p^n)^{\top} v_g^{n+1},\\
    x_p^{n+1} &=  x_p^{n} +  v_p^{n+1}\Delta t.
\vspace{-2pt}
\end{split}
\end{equation}

Contact between MPM and other modalities is typically handled on the background Eulerian grid. \textit{Grid-based contact model} directly zeros out the tangent component of grid velocity through $BC(\cdot)$ in \cref{eq:gridop}. Reaction force can be computed via momentum change $F=\Delta p / \Delta t$.
However, the grid nature makes it challenging to accurately represent and handle intricate geometries and moving boundaries. The method leads to severe penetration when coupling with delicate mesh-based objects.
Although scaling up grid resolution can alleviate the problem, it is typically not feasible in robotic simulation due to efficiency constraints.

Gu \textit{et al.} \cite{gu2023maniskill2} find that \textit{particle-based contact model} reduces penetration.
Specifically, the method approximates the contact surface as a spring: as a particle crosses the contact boundary with distance $d>0$, a force $F=-kd$ is exerted to push it back.
Essentially, the penalty force also works as an indirect constraint for grid velocity by adding a momentum term $F\Delta t$ to \cref{eq:p2g}, but provides more fine-grained control.
Particle-based contact model brings a problem on how to choose $k$. With a small $k$, penetration cannot be effectively alleviated. On the other hand, increasing $k$ will lead particles to rebound unnaturally near the boundary. Alleviating one of the artifacts makes the other worse.

\subsection{Forecast-based Contact Model}
We formulate the goal as reducing the number of penetrations at the end of each simulation loop. Since particle positions $x_p^{n+1}$ are advected based on velocities, we focus on how to set constraints on grid velocities to obtain $v_p^{n+1}$ that can achieve the goal. 
The idea of looking ahead and then adjusting grid velocity gives our method the name \textit{forecast-based contact model}.
It also provides an intuition why the method has better performance in reducing penetration.

We use $W$ as short for $W(x_p^n)$ from this section. Our method first takes a G2P transfer $v_{init}^{n+1} = W\hat v_g^{n+1}$.
Then we apply a boundary condition $BC_p(\cdot)$ on the particle velocities, which computes constrained velocities $v_{tgt}^{n+1}$ that are supposed to avoid penetration.
Given $v_{tgt}^{n+1}$, we transform our goal into an optimization problem:
\begin{equation}
\vspace{-2pt}
v_g^{n+1} = \argmin_{v_g} \| W^{\top} v_g - {v}^{n+1}_{tgt} \|^2.
\label{eq:forecast_objective}
\vspace{-2pt}
\end{equation}
The problem above can be tackled using iterative solvers like conjugate gradient method, whose computation overhead grows linearly with the number of iterations. However, we find that one-step gradient descent suffices in practice:
\begin{equation}
\vspace{-2pt}
v_g^{n+1} = \hat{v}_g^{n+1} - \alpha W (W^{\top} v_g - v_{tgt}^{n+1}),
\label{eq:one_step_gd}
\vspace{-2pt}
\end{equation}
where $\alpha$ is a predefined step size. 
The extra computation cost is equivalent to interpolating a particle property to grid and transferring it back. Besides, the derivatives of \cref{eq:one_step_gd} is simple enough to be directly obtained through automatic differentiation methods.

\begin{figure} [t!]
\centering
    \includegraphics[width=1.0\linewidth]{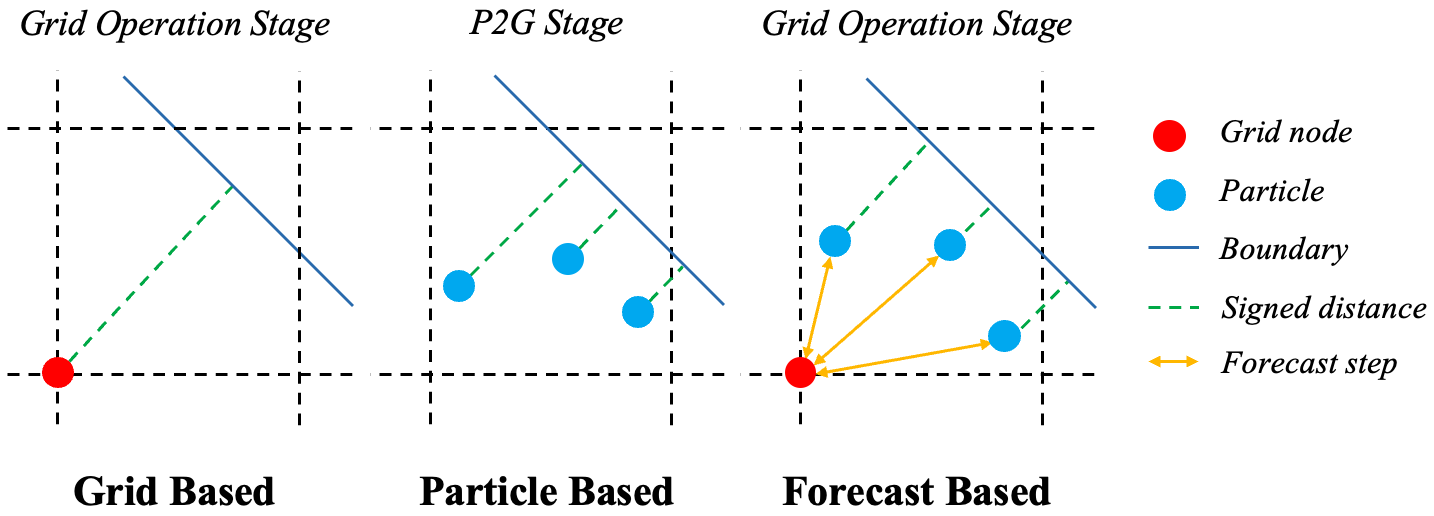}
    \caption{Illustration of the contact models. Grid-based model directly computes SDF on grid nodes. Particle-based model computes SDF on particles and applies penalty on them. Forecast-based model takes a forecast step to compute SDF on particles and then adjust grid velocity accordingly.
    }
    \label{fig:contact_illustration} 
\end{figure}

Next, we introduce the details of $BC_p(\cdot)$.
Given a particle with velocity $v_{in}$, we first check whether it is within the contact region by comparing signed distance $d$ with a threshold $\hat d$. 
If $d < \hat d$, the contact point is approximated with the nearest point on the boundary, whose velocity is $v_c$. 
For rigid bodies, $v_c$ is the linear velocity at the contact point.
For clothes, we find three vertices on the Lagrangian mesh that make up the contact surface, and compute weighted average of their velocities based on the barycentric coordinate of the contact point.
We then decompose the relative velocity $v_{rel} = v_{in} - v_{c}$ into a normal component $v_n$ and a tangential component $v_t$. To penalize the velocity, we drop $v_n$ and decay $v_t$ with friction:
\begin{equation}
\vspace{-2pt}
v_{out} = v_t\cdot \max\left(0, 1 - \mu {\| v_n \|} / {\| v_t \|}\right) + v_c,
\vspace{-2pt}
\end{equation}
where $\mu$ is the friction coefficient.
We also utilize two other techniques:
\textit{1)} Blend the original and modified velocities with a smoothness factor $s = \min\{ \exp(-\beta d), 1\}$, given by $v'_{out} = sv_{out} + (1-s)v_{in}$. $\beta$ is a predefined smoothness factor. The method reduces drastic state changes and improves gradient quality.
\textit{2)} Advect the particle with $v_{out}$, and if penetration happens, add a component to $v_{out}$ to ensure that the particles are moved to the nearest legal position.
Given velocity change, we can obtain impulse exerted on the particle, thereby computing the reaction force $F$.
The spatial force is accumulated on each rigid body link. For clothes, we distribute $F$ to three nodes of the nearest triangle based on the barycentric coordinate of the contact point.

\subsection{Comparisons between Contact Models}
As shown in \cref{fig:contact_illustration}, both grid-based and forecast-based contact models handle boundaries when solving grid velocities. The difference lies in that grid-based model computes SDF function on the static MPM grid nodes, while our method computes SDF function on Lagrangian particles that move with the material, allowing for more precise interaction with complex boundaries. Particle-based contact model also computes SDF on particles, and applies a penalty term during P2G transfer. However, the method is agnostic about how the penalty influences future particle states. In contrast, our method takes a \textit{forecast} step to see whether the particles will penetrate the boundary and utilizes the information to direct the update of grid velocity.

\subsection{Penetration Tracing for Soft-Cloth Coupling}

\begin{figure} [t!]
\centering
    \includegraphics[width=0.85\linewidth]{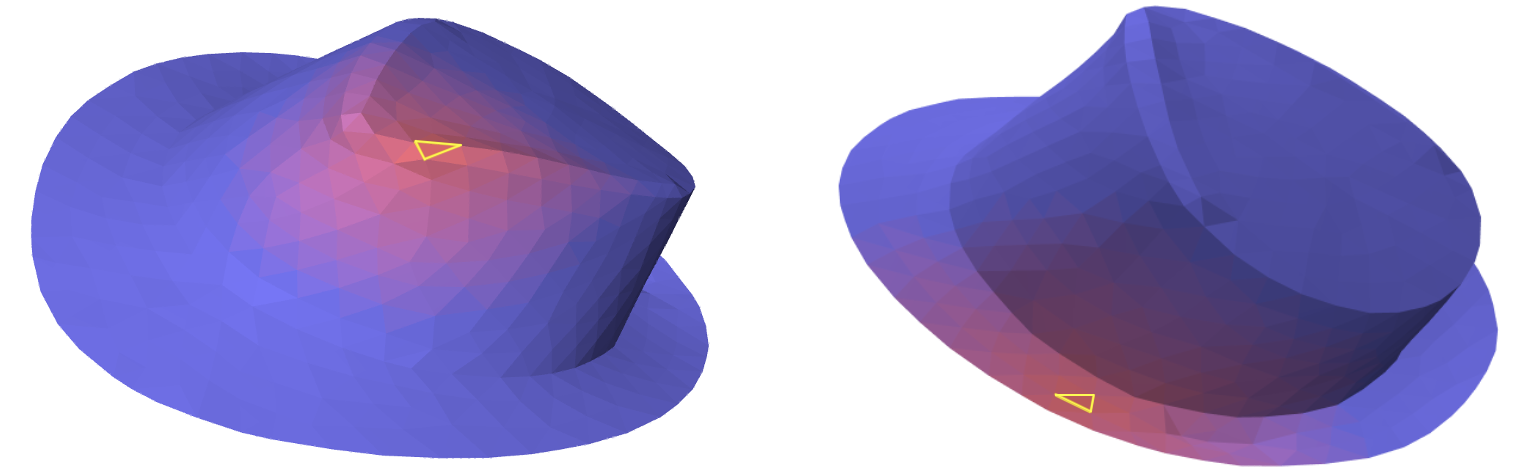}
    \caption{Given a target face (yellow edge), we search for the triangles in its neighboring area (red). If a particle is still in contact after system state changes, its nearest face should be within the neighboring area due to the small time interval.}
    \label{fig:confined_zone} 
\end{figure}

The contact models require signed distance to couple MPM soft bodies with objects represented as meshes. 
For rigid bodies, we pre-compute the signed distance field on compact grid points within the bounding boxes, and look up signed distance by interpolating the grid values during simulation.
However, the method is not applicable to soft-cloth coupling because meshes of clothes varies by time. Besides, such meshes are non-volumetric, which means that both sides can be considered as the outward surface. While it is feasible to manually define a positive side for some simple shapes like a square, the idea cannot generalize to special geometries such as a Möbius strip.

To fix the problem, we reconstruct SDF for cloth meshes by computing distance and sign separately.
We first search the particle-face pairs to find the nearest face for each particle, and compute \textit{absolute distances} accordingly. The process can be accelerated with spatial hashing. 
The \textit{sign} indicates whether a particle penetrated the mesh. Therefore, we assign a binary penetration state $z^{n}$ for each particle and trace it throughout the simulation process. Each time the position of particles or meshes change, we check whether the state change will lead to penetration. If a particle moves to the other side of the mesh, we update the penetration state at the next time step, given by $z^{n+1} = 1 - z^n$. Otherwise, $z^{n+1}$ inherits the value of $z^{n}$.

The complication lies in how to check whether penetration happens. We provide a method by utilizing the locality of motion in simulation.
The small time interval in physics simulation suggests that system state only changes a bit at each time. If a particle is in contact with face $i$ at time $n$ and face $j$ at time $n+1$, then we assume that faces $i$ and $j$ are within the neighboring area of each other. 
Based on this idea, we pre-compute the neighboring triangles around each face using breadth first search, and define a consistent orientation for faces in this local area.
Then we can directly check whether a particle resides on the same side of the area during simulation.
An illustration can be found in \cref{fig:confined_zone}.
% Given the sign and distance, we can determine whether and how a particle should be penalized using the contact models.
We discuss limitations of the algorithm in \cref{sec:conclusion}.

\section{Differentiable Simulation Design}

\begin{figure} [t!]
\centering
    \includegraphics[width=0.92\linewidth]{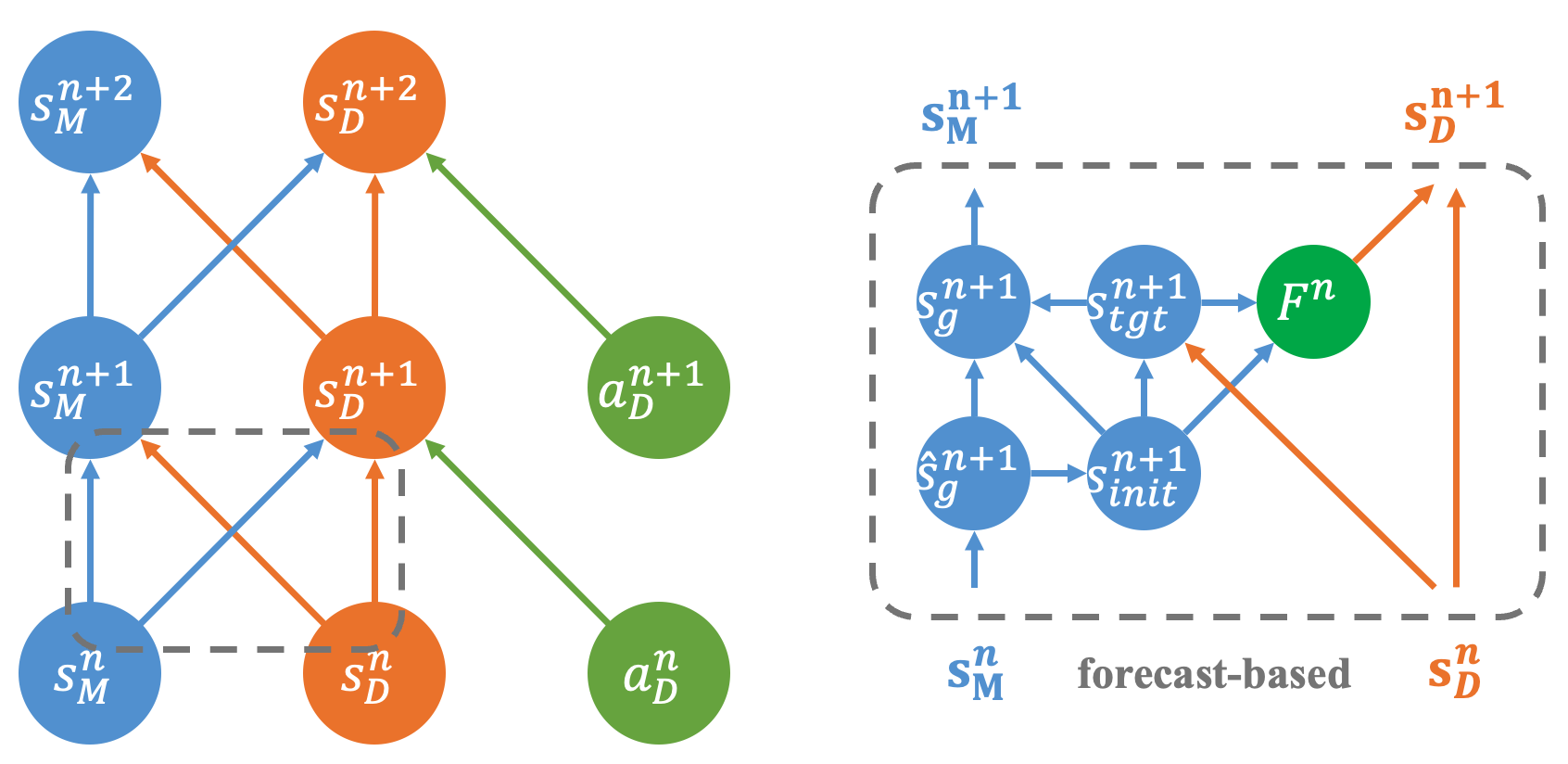}
    \caption{Computation graph for soft-rigid coupled system (the same in soft-cloth coupling). 
    \textit{Blue nodes}: state of MPM simulation. \textit{Orange nodes}: state of articulated rigid body simulation. \textit{Green nodes}: actions and external forces. 
    A line is connected between two nodes if and only if we directly use one state or action to compute the resulting state.
    The gray box illustrates the details of forecast-based contact model.
    }
    \label{fig:computation_graph} 
\end{figure}

\subsection{Forward Simulation}
Denote $s_M^{n}$ as state of MPM object at time $n$. The forward simulation for MPM is defined as $\mathcal{F}_M(s_M^n)$. 
$s_D^{n}$ and $a_D^{n}$ represent the state and action of the manipulator (articulated rigid bodies or clothes). The forward simulation for manipulator is $\mathcal{F}_D(s_D^n, a_D^n)$. While both simulations can be conducted independently, interactions between the two modalities cannot be directly simulated.
Therefore, we provide explicit coupling between these simulators as in \cite{gu2023maniskill2,han2019hybrid}.

Specifically, we modify the simulations as follows.
The MPM simulator takes poses and velocities of the manipulators as inputs. Based on the contact models, it updates MPM state and computes the reaction force on the manipulator. 
Manipulator simulation takes external force as an additional input, which is involved in the computation of the next state:
\begin{equation}
\vspace{-2pt}
\begin{split}
    s_M^{n+1}, F^{n+1} &= \mathcal{F}_M'(s_M^n, s_D^n), \\
    s_D^{n+1} &= \mathcal{F}_D'(s_D^n, a_D^n, F^{n+1}).
\end{split}
\vspace{-2pt}
\end{equation}

In a simulation loop, we execute $\mathcal{F}_M'$ first, with the manipulators as boundary conditions. 
Then we transfer the force back and execute $\mathcal{F}_D'$.
The computation graph is displayed in \cref{fig:computation_graph}.
We define state of the coupled system as $s^n = (s^n_M, s^n_D)$, and action as $a^n = a^n_D$. The forward simulation functions as:
\begin{equation}
    s^{n+1} = (s^{n+1}_M, s^{n+1}_D) = \mathcal{F}(s^n, a^n).
\end{equation}

\subsection{Backward Gradients}
The differentiable operations inside MPM simulation $\mathcal{F}'_M$ help us gather the Jacobian matrices $\frac{\partial s_M^{n+1}}{\partial s_M^{n}}$, $\frac{\partial s_M^{n+1}}{\partial s_D^{n}}$, $\frac{\partial F^{n+1}}{\partial s_M^{n}}$, $\frac{\partial F^{n+1}}{\partial s_D^{n}}$.
The manipulator simulation $\mathcal{F}'_D$ computes $\frac{\partial s_D^{n+1}}{\partial s_D^{n}}$, 
$\frac{\partial s_D^{n+1}}{\partial a^{n}}$, 
$\frac{\partial s_D^{n+1}}{\partial F^{n+1}}$.
Based on these Jacobian matrices, we apply the chain rule to obtain gradients for every system state at each time step.

$s_M^n$ is directly involved in the computation of $s_M^{n+1}$, and affects $s_D^{n+1}$ through the contact force $F^{n+1}$. Consequently, the gradient of $s_M^n$ has two components:
\begin{equation}
\frac{\partial \mathcal{L}}{\partial s_M^n} = 
\frac{\partial \mathcal{L}}{\partial s_M^{n+1}} 
\frac{\partial s_M^{n+1}}{\partial s_M^n} + 
\frac{\partial \mathcal{L}}{\partial s_D^{n+1}} 
\frac{\partial s_D^{n+1}}{\partial F^{n+1}}
\frac{\partial F^{n+1}}{\partial s_M^n}.
\end{equation}

$s_D^{n}$ influences $s_D^{n+1}$ in two ways.
In addition to the one inherent in $\mathcal{F}_D(s_D^n, a_D^n)$, $s_D^{n}$ affects the contact force $F^n$ between MPM and the manipulator, and the force in turn impacts on $s_D^{n+1}$. Another part of the gradient comes directly from $s_M^{n+1}$, which takes $s_D^n$ as the boundary:
% We calculate the gradients of $s_D^n$ as:
\begin{equation}
\frac{\partial \mathcal{L}}{\partial s_D^n} = 
\frac{\partial \mathcal{L}}{\partial s_D^{n+1}} 
(\frac{\partial s_D^{n+1}}{\partial s_D^n} + \frac{\partial s_D^{n+1}}{\partial F^{n+1}}
\frac{\partial F^{n+1}}{\partial s_D^{n}}) +
\frac{\partial \mathcal{L}}{\partial s_M^{n+1}} 
\frac{\partial s_M^{n+1}}{\partial s_D^{n}}.
\end{equation}

Finally, $a_D^n$ can be directly computed from $\frac{\partial \mathcal{L}}{\partial s_D^{n+1}}$ inside the manipulator simulation:
\begin{equation}
\frac{\partial \mathcal{L}}{\partial a_D^n} = 
\frac{\partial \mathcal{L}}{\partial s_D^{n+1}}
\frac{\partial s_D^{n+1}}{\partial a_D^{n}}.
\end{equation}

It is worth noting that $\frac{\partial s_M^{n+2}}{\partial a_D^n} = \frac{\partial s_M^{n+2}}{\partial s_D^{n+1}}\frac{\partial s_D^{n+1}}{\partial a_D^n}$. The gradients propagated from $s_M^{n+2}$ to $a_D^n$ explain why we can define losses on soft bodies to optimize the actions on the manipulators.

\section{Experiments}
In this section, we introduce our system implementation, and conduct quantitative and qualitative evaluations to answer the following questions:
1) Does forecast-based contact model produce more realistic simulation results? 2) Can penetration tracing effectively support soft-cloth coupling? 3) How accurate is the gradient information provided by the proposed two-way differentiable dynamics coupling?
In all the experiments, we maintain a CFL number below the critical threshold of 0.3 to ensure that the time integration scheme remains stable.

\begin{figure} [t!]
\centering
    \includegraphics[width=0.98\linewidth]{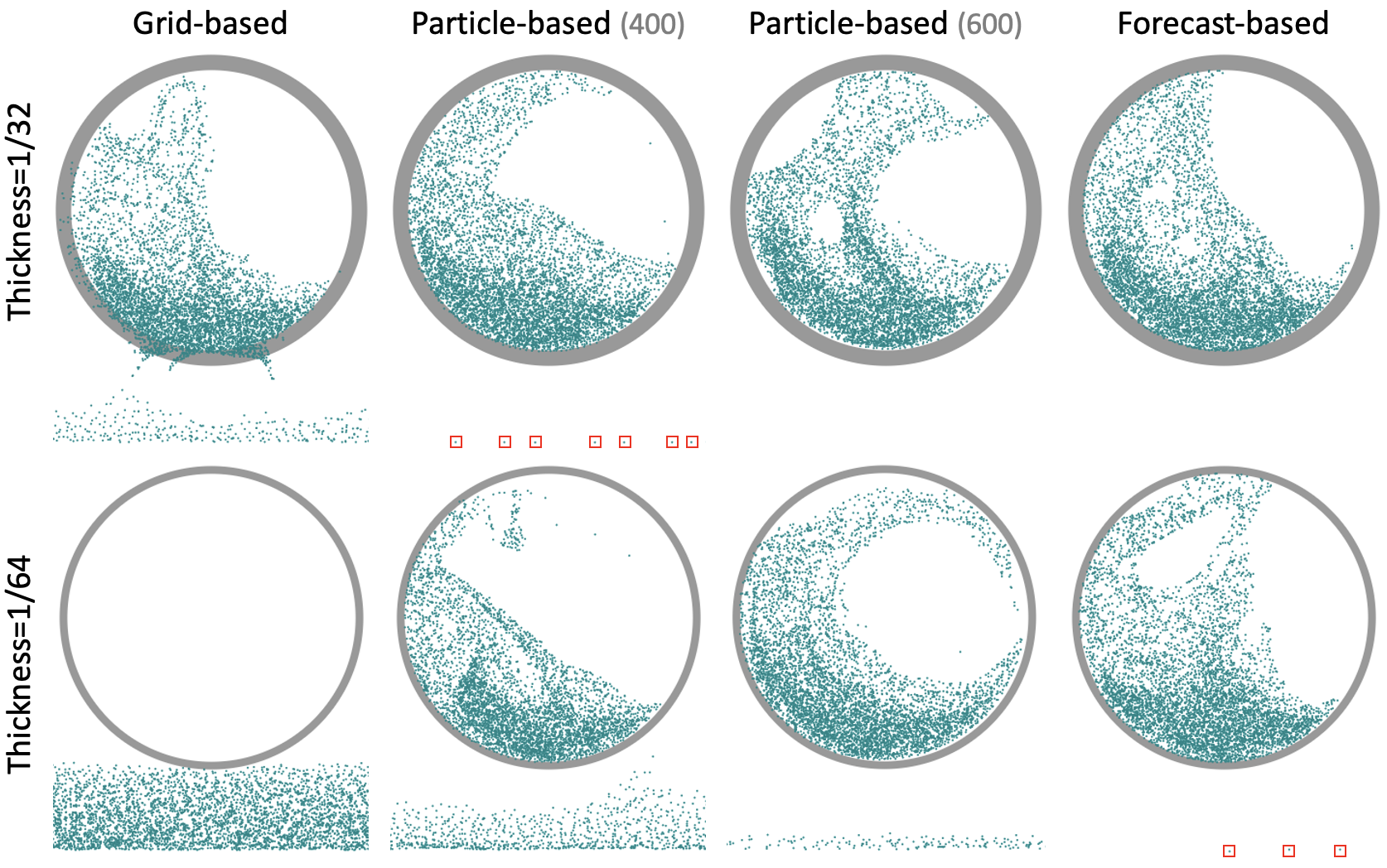}
    \caption{Shake a container with liquid. Grid-based model causes severe penetration. Particle-based model also faces the problem if the penalty coefficient $k$ is not high enough. Increasing $k$ alleviates penetration, but makes particles rebound more unnaturally near boundary. Forecast-based model results in least penetrations without bringing obvious artifacts.}
    \label{fig:contact_benchmark} 
\end{figure}

\subsection{System Implementation}
We implement a differentiable soft body simulator using DiffTaichi \cite{hu2019difftaichi}. Corotated constitutive models are used to simulate fluid, elastic and plastic materials. We refer readers to FluidLab \cite{du2012fluid} for more details about simulating different materials using MPM. 
We adopt two differentiable physics engines, Jade \cite{gang2023jade} and DiffClothAI \cite{yu2023diffclothai}, for the simulation of articulated rigid bodies and clothes, respectively.
In each simulation loop of the coupled system, we take actions, manipulator states, and MPM states as input, calculate contact in the soft body simulator, and transfer external force to the rigid body / clothes simulator. 
MPM typically requires a smaller time interval than the other two simulators. In this case, we take several MPM substeps and average the force on the manipulator.
Gradients are calculated accordingly.

\subsection{Checking Forecast-based Contact Model}

\begin{figure*} [t!]
    \centering
    \subfloat[Pour water\label{fig:demo_pour}]{
        \includegraphics[width=0.32\linewidth]{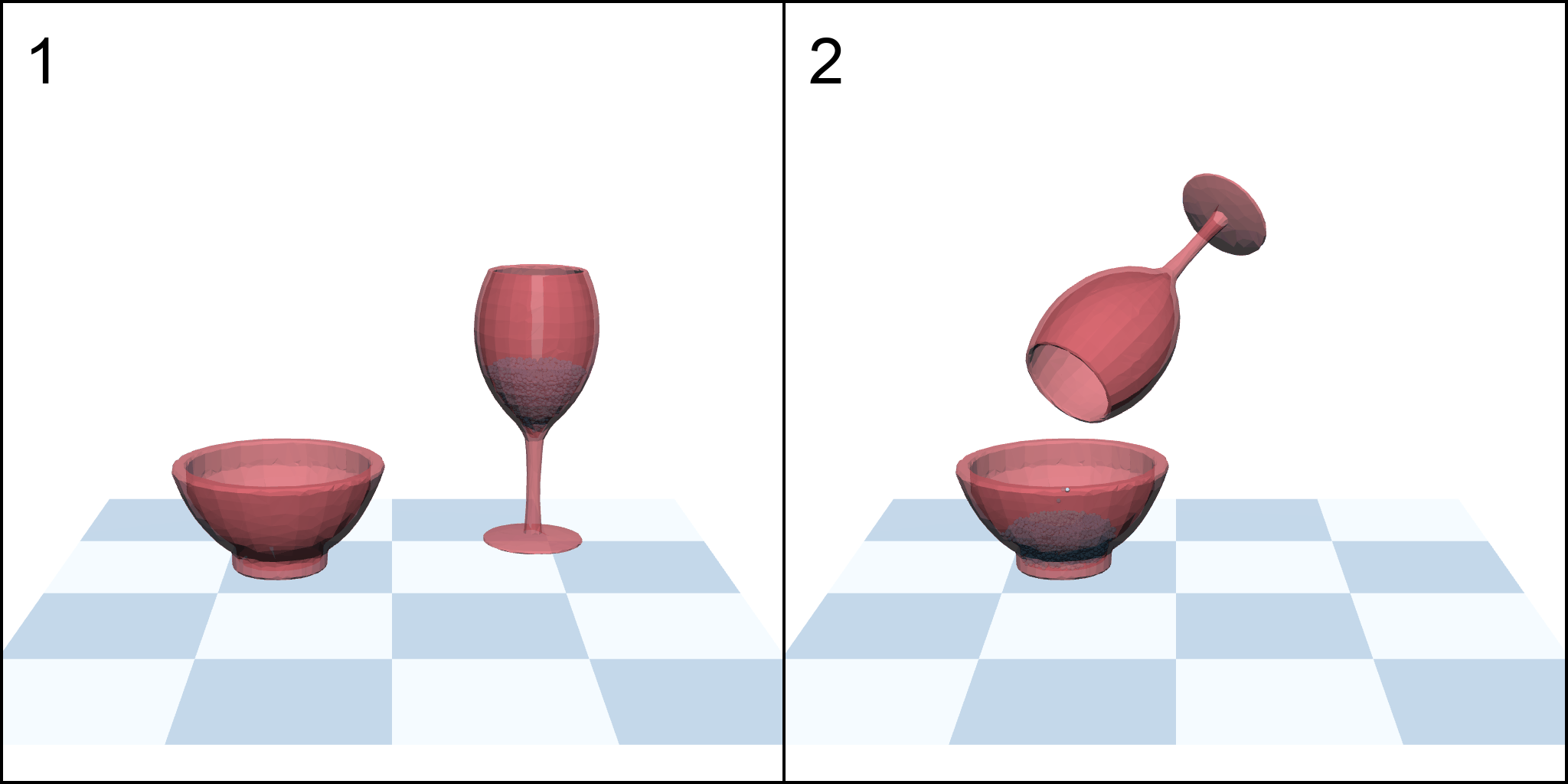}
    }
    \subfloat[Pour water (Franka)\label{fig:demo_pour_franka}]{
        \includegraphics[width=0.32\linewidth]{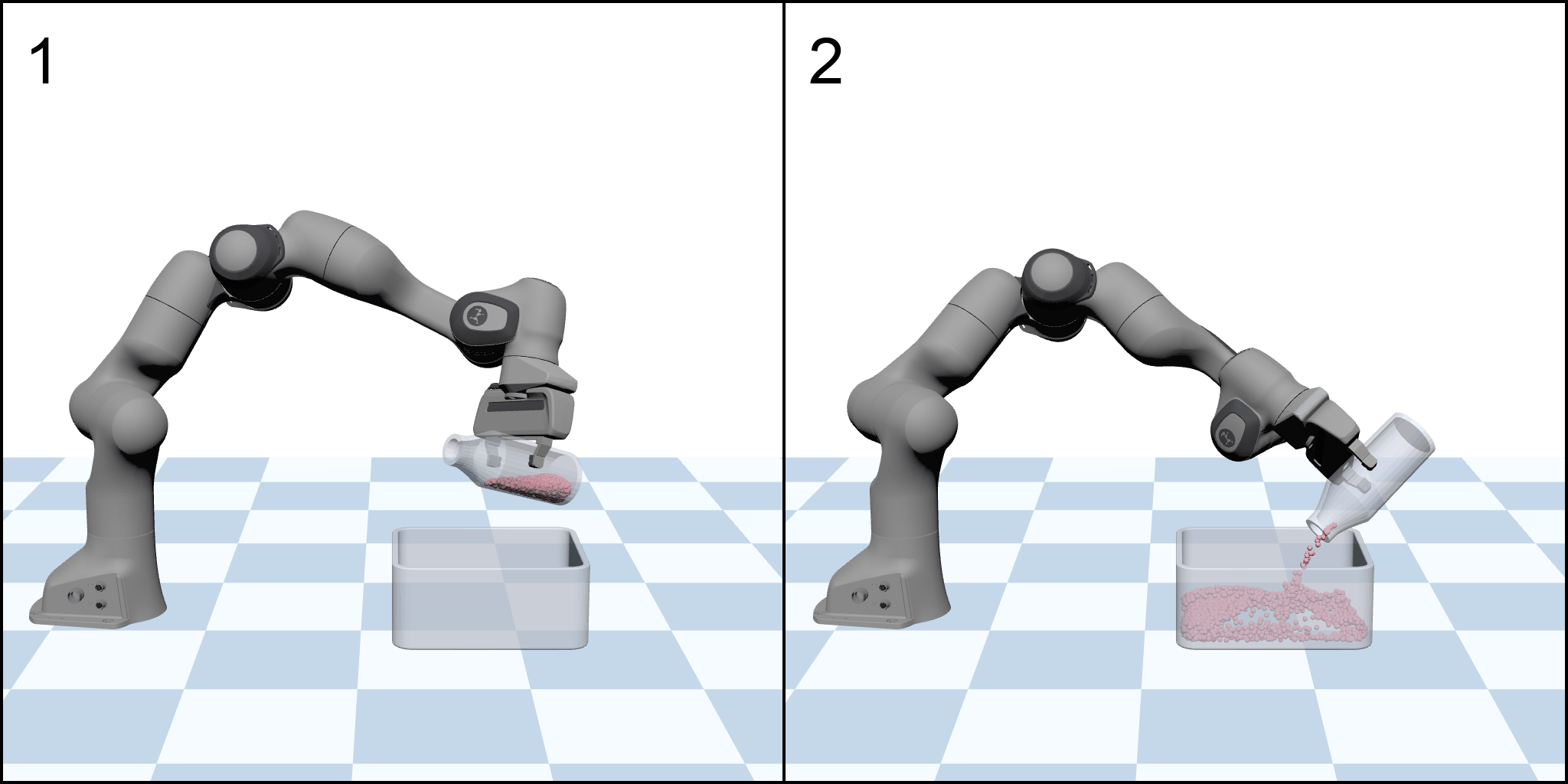}
    }
    \subfloat[Squeeze plasticine \label{fig:demo_grip}]{
        \includegraphics[width=0.32\linewidth]{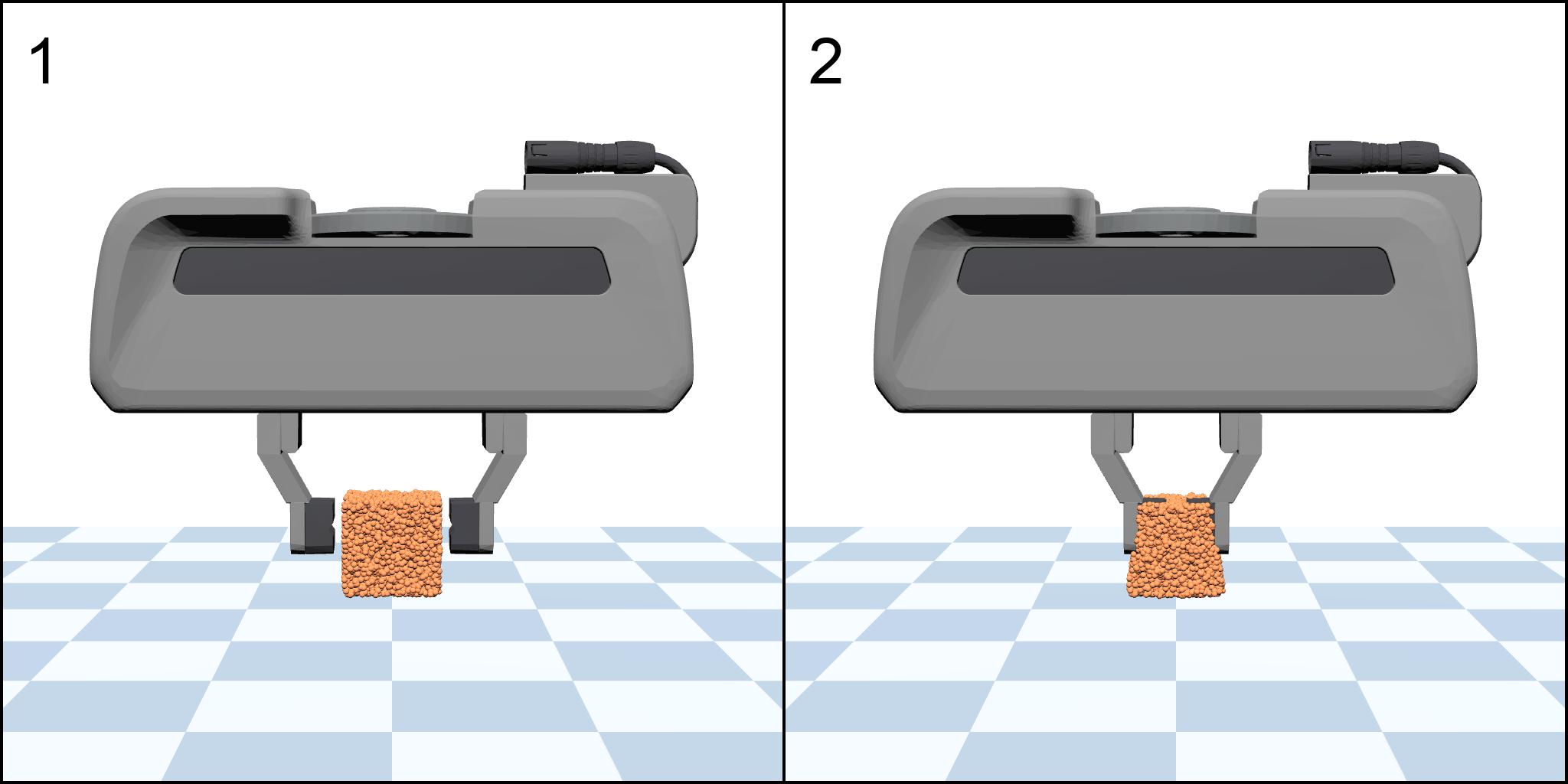}
    }
    \\
    \vspace{-0.5em}
    \subfloat[Pull door\label{fig:demo_door}]{
        \includegraphics[width=0.32\linewidth]{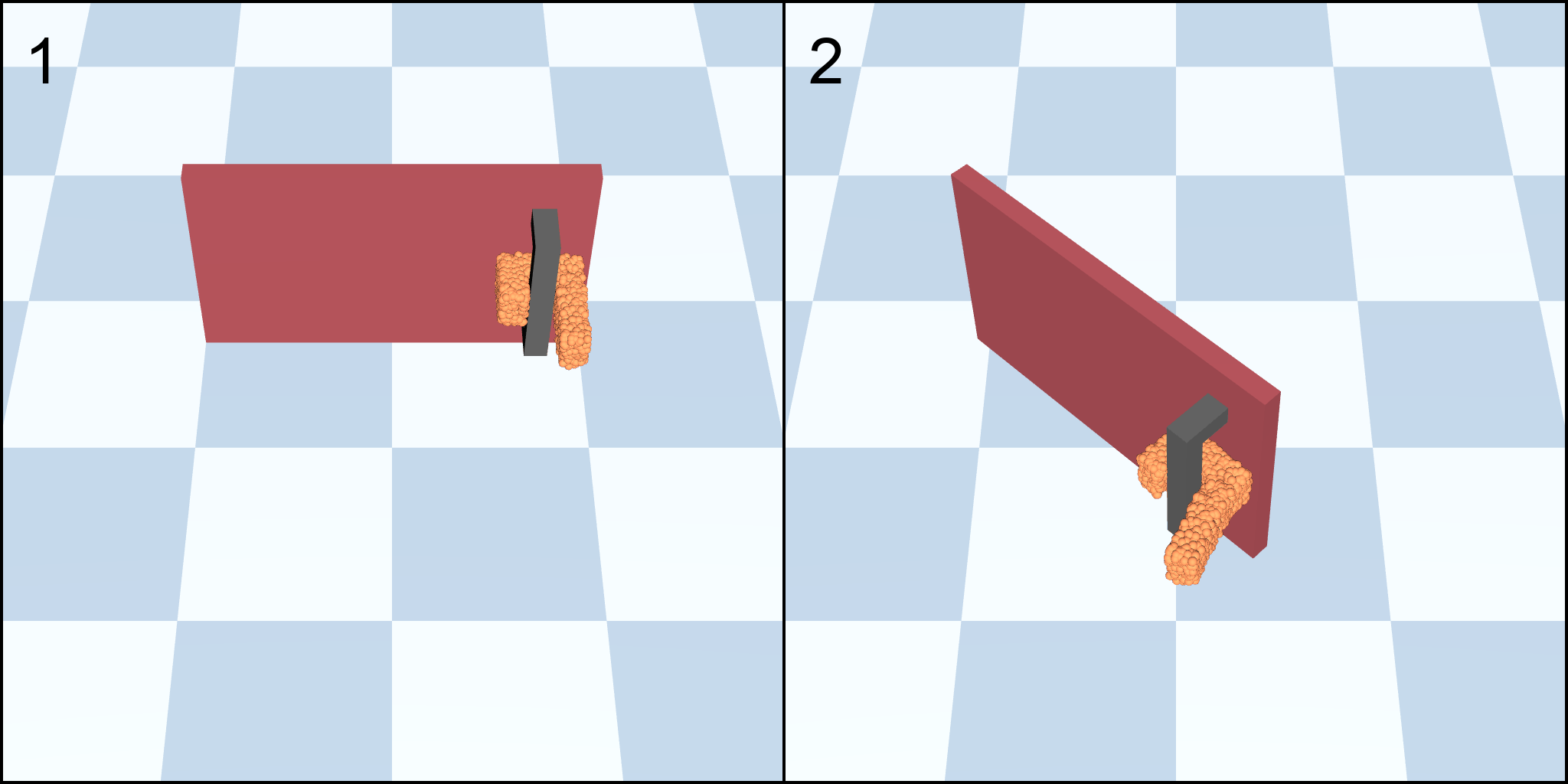}
    }
    \subfloat[Make taco\label{fig:demo_taco}]{
        \includegraphics[width=0.32\linewidth]{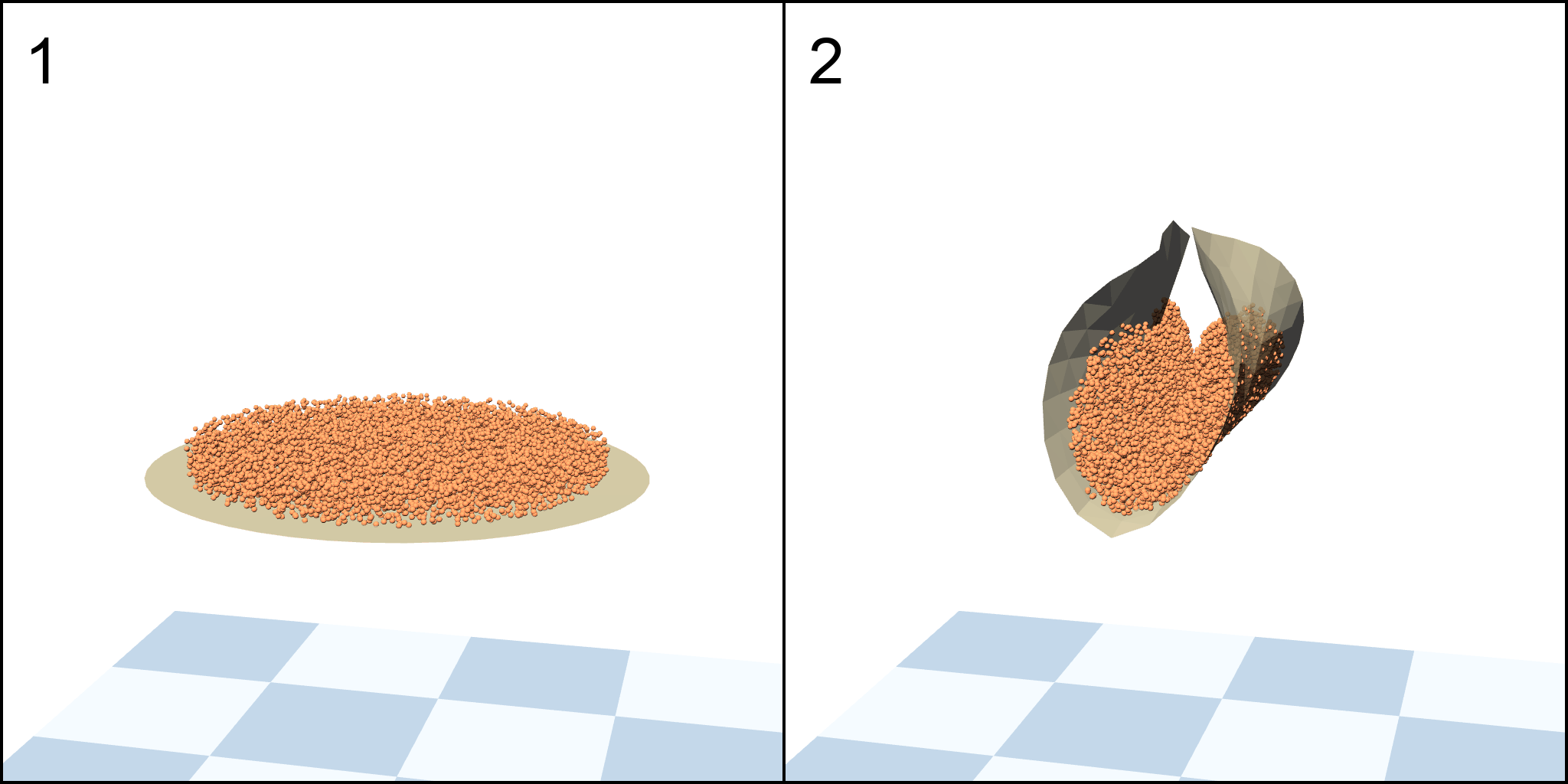}
    }
    \subfloat[Push towel\label{fig:demo_hit}]{
        \includegraphics[width=0.32\linewidth]{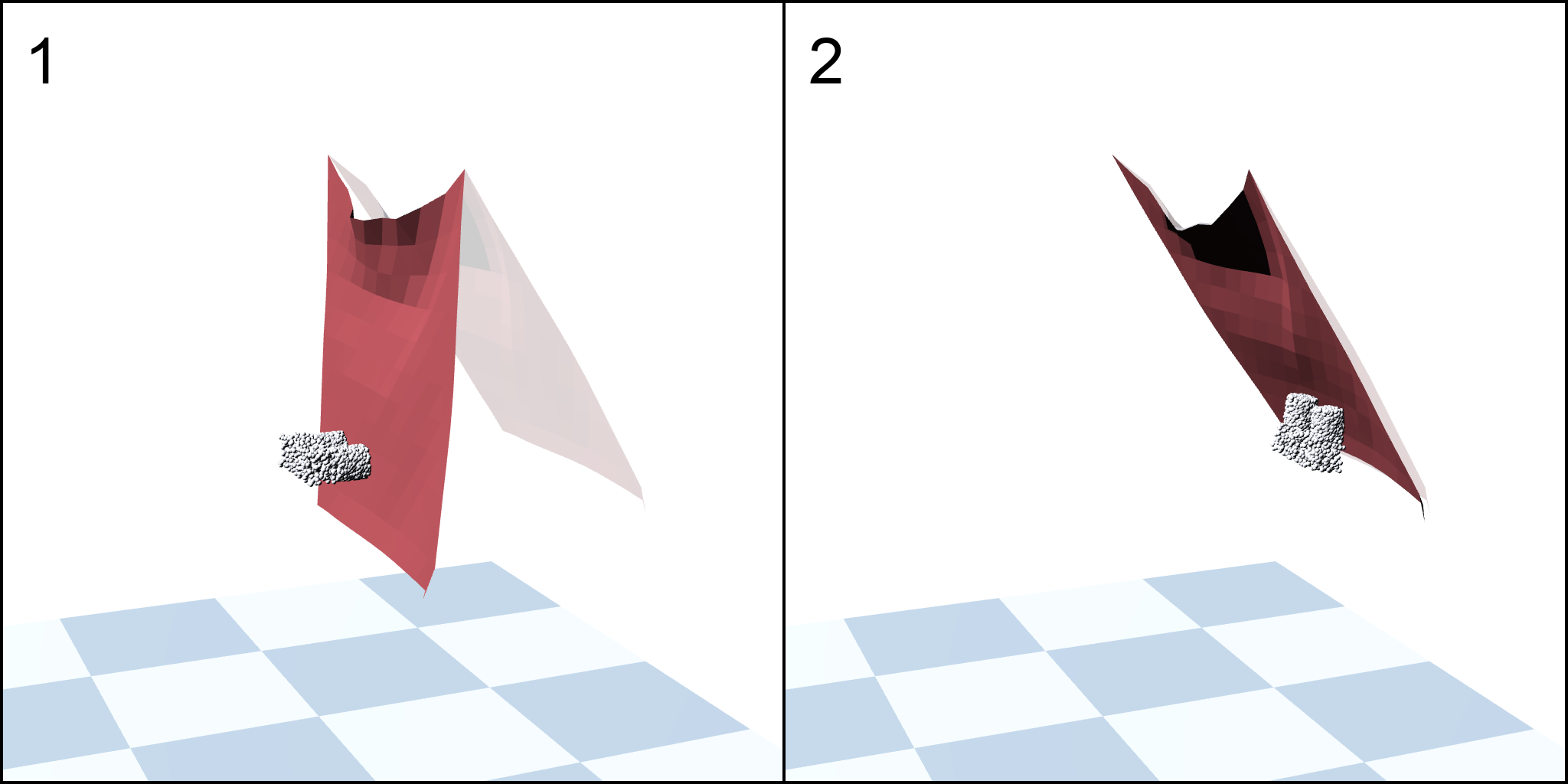}
    }
    \\
    \subfloat[Loss curves\label{fig:losses}]{
        \includegraphics[width=0.995\linewidth]{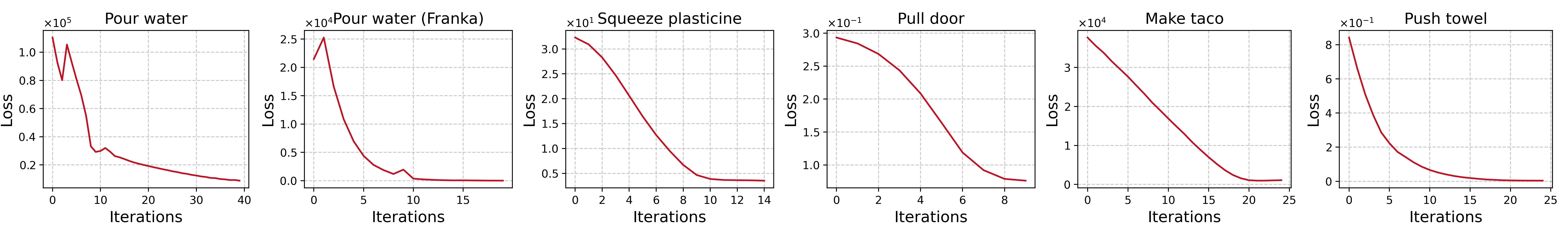}
    }

    \setlength{\abovecaptionskip}{0.2em}
    \caption{Experiment results of coupled differentiable simulation.
    \textit{(a)} Control glass to pour the liquid into a bowl.
    \textit{(b)} Control Franka arm to pour the liquid into a tank from a bottle.
    \textit{(c)} Control Franka gripper to squeeze the plasticine into target shape.
    \textit{(d)} Control 2 selected points on the tortilla to fold the taco into target shape.
    \textit{(e)} Control soft gripper to pull the door into target angle.
    \textit{(f)} Control soft gripper to push the towel into target pose.
    % We define targets on particles and apply actions on rigid bodies / clothes to ensure that gradients correctly propagates between different materials.
    In all tasks, their respective training loss curves are plotted on the right. Full trajectories are displayed in our video.
    }
    \label{fig:demos} 
\end{figure*}

We first conduct a quantitative experiment to compare forecast-based contact model with the baselines (grid-based and particle-based models). 
Specifically, we initialize the 2D scene with fluid inside a circular rigid container, which has radius $r$ and thickness $d$.
% which has a inner surface of radius $r_1$ and an outer surface of radius $r_2$.
The container shakes left and right at a constant speed, and we record the number of particles penetrating the container.
% We denote distance to the center of circle as $d$. 
% Three metrics are introduced: \textit{1)} the number of particles inside the container wall ($r_1 < d \leq r_2$); \textit{2)} the number of particles that penetrate the outer surface of the container ($d > r_2$); \textit{3)} computation time.
% Since the contact models are not penetration-free, we provide metric 1 as reference and focus on more metric 2.

We choose step size $\alpha=0.2$ for the one-step gradient descent in forecast-based contact model, and the objective in \cref{eq:forecast_objective} decreases by 83.1\% on average. 
As shown in \cref{tab:contact_model_stat}, forecast-based model achieves the best performance in reducing penetration.
The computation efficiency lags behind baselines as we introduce additional P2G and G2P transfers.
Another empirical observation is that while increasing the penalty coefficient $k$ in particle-based contact model can alleviate penetration, it also exacerbates unnatural rebound near the boundary. 
Meanwhile, our method does not have this problem.
Visualization for the last frame is provided in \cref{fig:contact_benchmark}.
We also conduct a qualitative experiment: pour liquid into a thin glass with high velocity.
Results shown in \cref{fig:contact_models} are consistent with our quantitative experiment.

\subsection{Checking Penetration Tracing}
We drag the four corners of a towel to squash a plasticine. The control points are moved down from their initial positions at a constant speed. 
We add and remove the penetration tracing algorithm to demonstrate the effectiveness of our method in soft-cloth coupling.
The results are displayed in \cref{fig:penetration_tracing}.
We find that lacking the penetration information causes the towel to go though the plasticine directly.
However, with penetration tracing, the plasticine is squashed and the towel deforms due to the reaction force.
We conduct the comparison on particle-based contact model. With forecast-based model, the towel cannot pass the plasticine even after removing the tracing algorithm in this case.

\begin{table} [t!]
    \centering
    \setlength\tabcolsep{3.8pt}
    \renewcommand{\arraystretch}{1.15}
    \begin{tabular}{l|c|c|c|c}
    \toprule[1.2pt]
        \textbf{Contact Model} & \textbf{Thickness} & \textbf{\#Penetration}$\downarrow$ & \makecell{\textbf{Unnatural}\\\textbf{Rebound}} & \textbf{Time (s)} \\\hline
        \multirow{2}{*}{\textbf{Grid}} & 1/32 & 449 & \xmark & 4.90 \\
         & 1/64 & 5998 & \xmark & 4.81 \\\hline
        \multirow{2}{*}{\textbf{Particle (k=400)}} & 1/32 & 12 & \xmark & 4.45 \\
         & 1/64 & 1512 & \xmark & 4.71 \\\hline
        \multirow{2}{*}{\textbf{Particle (k=600)}} & 1/32 & \textbf{0} & \cmark & 4.33 \\
         & 1/64 & 217 & \cmark & 4.68 \\\hline
        \multirow{2}{*}{\textbf{Forecast (Ours)}} & 1/32 & \textbf{0} & \xmark & 5.24 \\
         & 1/64 & \textbf{3} & \xmark & 5.15 \\
    \bottomrule[1.2pt]
    \end{tabular}
    \caption{
        Experiment results of shaking containers of different thickness $d$ with liquid for 1100 frames. 
        % Experiment results of 1100 frames of liquid shaking in containers of different thickness $d$.
        \#Penetration denotes the number of particles outside the container in the final frame. 
        Unnatural rebound occurrences are recorded. 
        Simulation time is compared on an NVIDIA Tesla T4 GPU.
    }
    \label{tab:contact_model_stat}
\end{table}

\subsection{Checking Two-way Differentiable Dynamics Coupling}
We check the quality of two-way differentiable dynamics coupling under 6 robotic manipulation tasks. 
% Detailed descriptions of each task are presented in \cref{tab:task_description}.
% While the first 3 tasks can be done in existing simulations using kinematic skeletons as controller, \textit{pull door}, \textit{make taco}, and \textit{push towel} can only be done in SoftMAC as dynamics of both modalities must be involved to fulfill the task.
The first three tasks can also be accomplished within current simulations that employ kinematic skeletons as controllers. 
However, \textit{pull door}, \textit{make taco}, and \textit{push towel} necessitates the utilization of SoftMAC as the dynamics of both modalities must be incorporated to successfully complete these tasks.

We formulate our problem as trajectory optimization: Given an action sequence $ a = (a_1,a_2,\cdots,a_T)$ and the resulting state sequence $ s = (s_{1}, s_2,\cdots, s_T)$, we define an objective function $f( a,  s)$ and find the optimal actions through
\begin{equation}
% \vspace{-2pt}
{a}^* = \argmin_{ a} f( a, s).
\label{eq:traj_opt}
% \vspace{-2pt}
\end{equation}
We use the gradients provided by SoftMAC and first-order optimizers (SGD for \textit{pouring water (franka)} and Adam \cite{kingma2014adam} for other tasks) to solve \cref{eq:traj_opt}. 
The number of variables to optimize ranges from 200 to 1150.
It is worth noting that for more complicated tasks, the trajectories are prone to get stuck in local optima. 
Initialization from teleportation or learned policies helps overcome the problem \cite{li2023dexdeform}, but is beyond the scope of our experiments.

\paragraph{Rigid2MPM} 
We evaluate soft-rigid coupling on three tasks.
In \textit{pour water}, we conduct 6 DoFs control over a glass to pour liquid into a bowl.
In \textit{pour water with Franka}, we conduct 7 DoFs control over the Franka arm to pour liquid from a bottle into a tank.
In \textit{squeeze plasticine}, we conduct 2 DoFs control over a gripper to squeeze the 16200 DoFs plasticine into target shape.
Losses are computed as the Chamfer distance between current and target positions of the particles.
In the pouring tasks, two additional loss terms are calculated on poses and velocities of the source container to penalize it from colliding with the target container.

\paragraph{Cloth2MPM} 
We verify soft-cloth coupling with \textit{make taco}.
The tortilla and taco fillings are simulated as clothes and MPM particles, respectively.
We conduct 4 DoFs control over two selected points of the tortilla and fold the taco into target shape.
% with 200 steps (2000 MPM substeps).
The state space is combination of the 30000 DoFs taco fillings and the 651 DoFs tortilla.
Loss is computed on taco fillings as the Chamfer distance between current and target configurations.
To avoid overly stretching the tortilla, we control the particles in the perpendicular plane through circle center and project the control points back to a confined region after each epoch.
% We optimize the 4 DoFs action space to minimize the objective using Adam as optimizer.

\paragraph{MPM2Rigid/Cloth} We check two-way differentiable coupling by controlling soft grippers to manipulate articulated rigid bodies or clothes. 
We control MPM by exerting an impulse on selected particles in P2G transfer. 
In \textit{pull door}, we control the elastic gripper to pull the door into target angle.
In \textit{push towel}, we control the elastic gripper to push the 432 DoFs towel into target pose.
We computed losses on the state of rigid bodies / clothes and optimize the control over soft grippers.

The pour water and make taco tasks are optimized for 40 and 25 iterations respectively, costing 21.67s and 20.45s for each iteration on average on an NVIDIA Tesla T4 GPU.
Experiment results reported in \cref{fig:demos} demonstrate the correctness and effectiveness of our differentiable simulation. 

% \section{Limitations}
% \label{sec:limitation}

% \paragraph{Penetration tracing for soft-cloth coupling}
% In this work we operate under the assumption that cloth meshes are manifold and not pressed together.
% The assumption is reasonable in the underlying cloth simulator \cite{yu2023diffclothai} that uses Incremental
% Potential Contact \cite{li2020incremental} to guarantee a gap between meshes.
% % Finding a general-purpose tracing algorithm for MPM particles under stress cases (\textit{e.g.}, multiple layers of fabric are stacked) is challenging and remains an open problem.
% Addressing penetration tracing challenges in complex fabric arrangements remains an open problem and requires further exploration for a comprehensive solution.

% % The algorithm is limited to manifold geometry and scenarios where clothes meshes spread out.
% % Counter examples can be found under stress cases. 
% % For instance, if multiple layers of fabric are stacked, a particle might penetrate a face not neighboring to the previous contact face.
% % Finding a general-purpose tracing algorithm for MPM particles is challenging and remains an open problem.

% \paragraph{Computation efficiency}
% The MPM simulator in SoftMAC is developed using Taichi to support massive parallelism on GPUs. 
% However, the rigid body and cloth simulators run on CPUs. Frequent states and gradients transfer between CPUs and GPUs brings extra computation overhead and impacts overall performance.
% We leave the development of GPU-based differentiable rigid body and cloth simulators with necessary interfaces as future work.

\section{Conclusion and Future Work}
\label{sec:conclusion}
% \looseness=-1
In this work, we introduce SoftMAC, a differentiable simulation that couples soft bodies with articulated rigid bodies and clothes. 
We simulate soft bodies with material point method, and provide a novel contact model that reduces artifacts like penetration and unnatural rebound.
To couple soft body particles with deformable and non-volumetric clothes meshes, we propose a method to reconstruct signed distance field in local area. 
We also utilize a two-way differentiable dynamics coupling mechanism to unify the simulation of different materials.
Comprehensive experiments are conducted in robotic manipulation scenarios and demonstrate the correctness and effectiveness of the differentiable simulation.
% In a broader sense, SoftMAC is the first differentiable robotic simulator that supports bidirectional soft-rigid and soft-cloth coupling. Our work pushes forward the development of universal differentiable physics simulation.

One limitation of this work lies in the assumption that cloth meshes are manifold and not pressed together for penetration tracing.
The assumption is grounded in the underlying cloth simulator \cite{yu2023diffclothai} that uses Incremental Potential Contact \cite{li2020incremental} to guarantee a gap between meshes.
However, addressing penetration tracing challenges in complex fabric arrangements remains an open problem and requires further exploration for a comprehensive solution.
Another drawback is rooted in computational efficiency. 
The MPM simulator in SoftMAC is developed using Taichi to support massive parallelism on GPUs. 
However, the rigid body and cloth simulators run on CPUs. 
The rate of state and gradient transfer between CPUs and GPUs is restricted by bandwidth and impacts overall performance.
We leave the development of GPU-based differentiable rigid body and cloth simulators with necessary interfaces as future work.

%%%%%%%%%%%%%%%%%%%%%%%%%%%%%%%%%%%%%%%%%%%%%%%%%%%%%%%%%%%%%%%%%%%%%%%%%%%%%%%%

{
\bibliographystyle{IEEEtran}
\bibliography{ref}
}

\end{document}